\documentclass[conference]{IEEEtran}
\IEEEoverridecommandlockouts
\usepackage{cite}
\usepackage{amsmath,amssymb,amsfonts}
\usepackage{graphicx}
\usepackage{textcomp}
\usepackage{xcolor}
\usepackage{amsmath}
\newcommand{\st}{\text{subject to}}
\newcommand{\E}{\mathbb{E}}

\usepackage{matlab-prettifier}
\usepackage{color}
\usepackage[shortlabels]{enumitem}
\usepackage{algorithm}
\usepackage[noend]{algpseudocode}
\makeatletter
\def\BState{\State\hskip-\ALG@thistlm}
\makeatother
\def\BibTeX{{\rm B\kern-.05em{\sc i\kern-.025em b}\kern-.08em
    T\kern-.1667em\lower.7ex\hbox{E}\kern-.125emX}}
\usepackage{subfigure}

\begin{document}

\title{Cognitive Level-$k$ Meta-Learning for Safe and Pedestrian-Aware  Autonomous Driving \\
\thanks{Both authors are with the Department of Electrical and Computer Engineering, Tandon School of Engineering, New York University; E-mail: \{hl4155,qz494\}@nyu.edu}
}

\author{\IEEEauthorblockN{Haozhe~Lei and Quanyan~Zhu}

}

\maketitle

\begin{abstract}
The potential market for modern self-driving cars is enormous, as they are developing remarkably rapidly. At the same time, however, accidents of pedestrian fatalities caused by autonomous driving have been recorded in the case of street crossing. To ensure traffic safety in self-driving environments and respond to vehicle-human interaction challenges such as jaywalking, we propose Level-$k$ Meta Reinforcement Learning (LK-MRL) algorithm. It takes into account the cognitive hierarchy of pedestrian responses and enables self-driving vehicles to adapt to various human behaviors. 
We evaluate the algorithm in two cognitive confrontation hierarchy scenarios in an urban traffic simulator and illustrate its role in ensuring road safety by demonstrating its capability of conjectural and higher-level reasoning.
\end{abstract}


\section{Introduction}
\begin{figure}[h]
    \centering
   \includegraphics[width=0.49\textwidth]{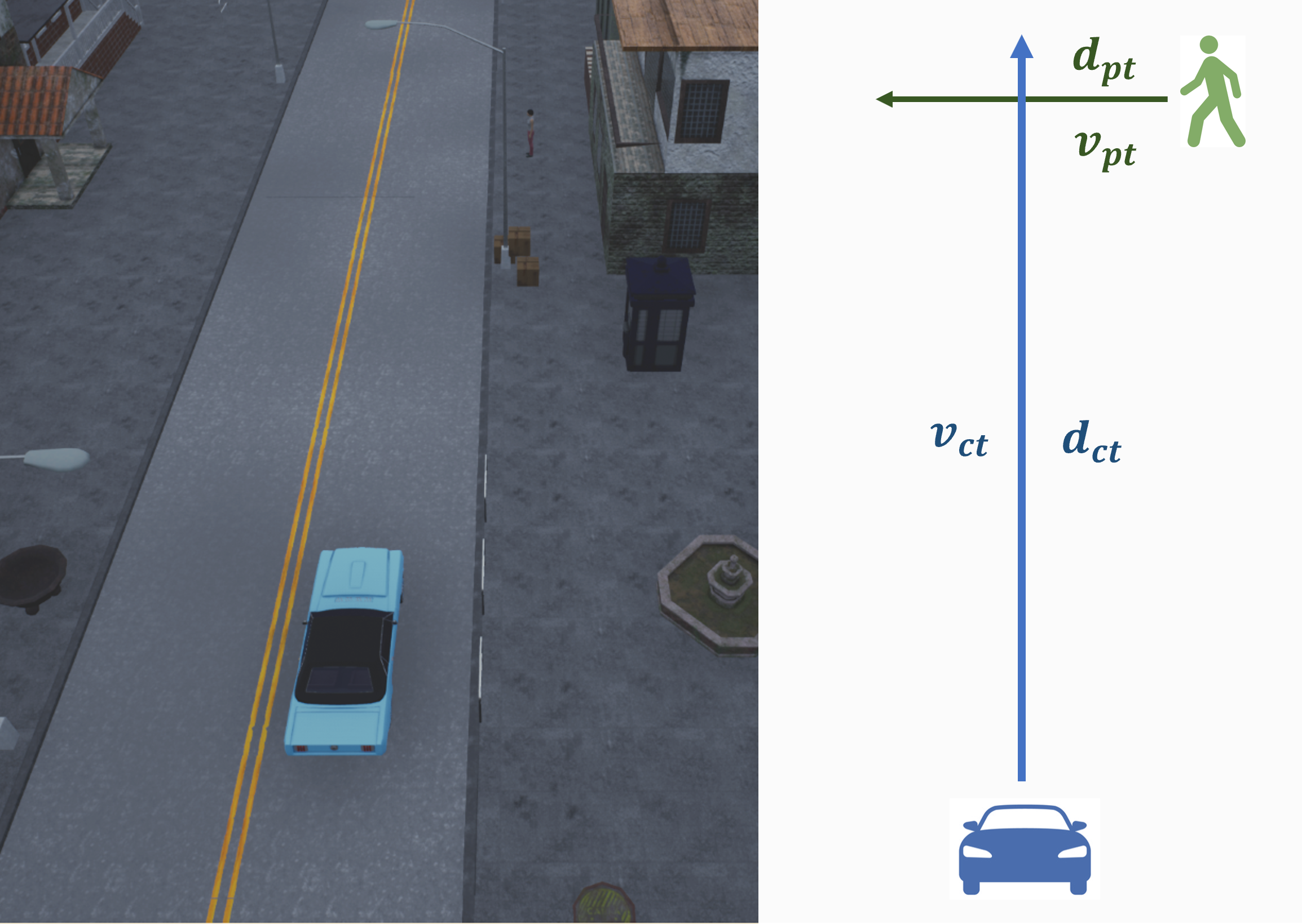}
    \caption{An illustration of the jaywalking scenario. In this, we create a pedestrian 30 meters in front of the car on the urban sidewalk road. The sensors will observe the velocities $v_{ct}, v_{pt}$ and their distances to the destination $d_{ct},d_{pt}$ of the pedestrian and the vehicle. The vehicle needs to reach its destination in a short period of time without colliding with pedestrians.}·
    \label{fig:scenario_true}
\end{figure}

One challenge for self-driving cars is their interactions with vehicles as well as pedestrians in urban environments. The unpredictability of pedestrian behaviors at intersections can lead to a high rate of accidents. The first pedestrian fatality caused by autonomous vehicles was reported in 2018 when a self-driving Uber vehicle struck a woman crossing an intersection in Tempe, Arizona, in the nighttime \cite{wakabayashi2018self}. At the same time, many studies have examined the risks of accidents (e.g., \cite{SHI2018346}) and post-accident measures (e.g., \cite{6850749} ). It is much more appropriate to avert accidents than to remedy them afterward. To be more precise, there is a need for creating machine intelligence that allows autonomous vehicles to control the car and adapt to different pedestrian behaviors to prevent accidents. There is a quintessential scenario for vehicle-human interactions in urban traffic is the case where the self-driving car handles jaywalking shown in figure \ref{fig:scenario_true}.

Reinforcement learning is central to the development of end-to-end control algorithms. The extension to the Multi-Agent Reinforcement Learning (MARL) algorithms~\cite{multi-agent-system-survey} is suitable for the modeling of vehicle-human interactions and be thought of as a solution for preventing traffic accidents. However, there is a crucial problem with classical MARL in this area stemming from the fundamental differences between machines and human agents. Earlier studies, such as \cite{Pedestrian-rich,Model_Uncertainty_Estimates,reverse_rl}, have created fixed pedestrian models when developing learning-based control algorithms. They have not sufficiently captured the variability in pedestrian behaviors, including reaction time, cognitive capabilities, and their dynamic response to the environment. Therefore, challenges remain in complex urban environments where vehicles need to interact with different pedestrians. Human agents have limited cognitive and reasoning abilities. There is a need to differentiate the computational and reasoning capabilities between pedestrians and vehicles.

Level-$k$ thinking \cite{level-k}, as a cognitive hierarchy framework, can provide a behavioral approach to capture bounded human reasoning processes in strategic interactions. It has shown promising accuracy in the prediction of human behaviors in contrast to analytical methods \cite{human_behavior}. 
In this work, we incorporate the cognitive hierarchy framework into MARL by viewing pedestrians and vehicles as different cognitive level-$k$ agents who respond to others with bounded rationality. To capture the variability in human behaviors, we leverage the recent advances in meta reinforcement learning (meta-RL) algorithms, such as Model-Agnostic Meta-Learning (MAML)~\cite{pmlr-v70-finn17a} and Conjectural Online Lookahead Adaptation (COLA)~\cite{COLA}, to augment the MARL with the ability of fast adaptation to heterogeneous pedestrian behaviors and changing environments. 

The contribution of this paper is the algorithm that allows self-driving vehicles to adapt to its conjecture of pedestrians' behavior by cognitive hierarchy. The algorithm consolidates level-$k$ thinking into MAML based on the current behavior of pedestrians and trained policy to realize online adaptation.

\subsection{Related works}

Many earlier works have relied on imitation methods to generate the data and have assumed that pedestrians have similar response patterns. Recent works in  \cite{Pedestrian-rich, Model_Uncertainty_Estimates} have trained models using robot hardware data and simulation data from static pedestrian agents. 
Since the models are trained offline, there is a need for capability to adapt to diverse human behaviors. 
%
Level-$k$ thinking, as a hierarchical model, has very strong interpretability. The consolidation of level-$k$ thinking and RL has become widely studied in strategic decision-making. Authors of \cite{bahaviour_qlearning_levelk,klevel-lighting} have modeled behavioral predictions of drivers in highway driving scenarios by using level-$k$ thinking and deep $Q$-learning (DQN). In addition, \cite{bounded_marl} has introduced generalized recursive reasoning (GR2) as a novel framework to model agents with different hierarchical levels of rationality to solve MARL problems, while \cite{GCH} establishes a generalized cognitive hierarchy (GCH) model which assumes that level-$k$ best-responds to all lower levels. However, high-dimensionality  of the learning task remains a challenge together with the vehicle-human interactions. 

\subsection{Paper Organization}
Section \ref{sec:model} will give the description of the model and the review of preliminary knowledge. Section \ref{sec:mechanism} will show how to obtain the model by solving alternate optimization problems, and section \ref{sec:simulation} will provide the simulating results to show the algorithm's evaluations. The paper is concluded in section \ref{sec:conclusion}.




\section{System Model and Preliminary}\label{sec:model}
This section will first introduce the meta-RL method we use in this work and the basic idea of level-$k$ thinking, respectively. Then, we will describe the model and question setting in this paper.

\subsection{Model Description}\label{sec:model_description}

Following the figure \ref{fig:scenario_true}, without loss of generality, we consider a model of two agents $i\in\mathcal{I}:=\{c,p\}$. One is the vehicle, denoted by $c$, and the other is the pedestrian, denoted by $p$. For modeling the variability in pedestrian behaviors, Each player is associated with a type live in a type-space $\Theta_i$. In particular, we let $\Theta_c:=\{0\}$ be a singleton set, indicating that the vehicle's type of certain, and $\Theta_p:=\{ 1, 2, 3\}$, indicating that there are three cognitive levels for pedestrians. Then $j\in \Theta_i$ would be the index of the type associated with agent $i$.

The two players interact dynamically with a $m$ dimensional sensor, and both have $n$ dimensional controller. Let $t\in \mathbb{N}$ be the time index and the state of the system be given by $s_t\in S\subseteq \mathbb{R}^{m}$ at time $t$. The state consists of each player's current and previous speeds $v_{ct}, v_{pt} \in \mathbb{R}$, distance to their endpoints $d_{ct},d_{pt} \in \mathbb{R}$, and actions $a_{c,t}\in \mathcal{A}_c\subseteq \mathbb{R}^{n}, a_{p,t}\in \mathcal{A}_p\subseteq \mathbb{R}^{n}$. The full structure of state $s_t$ is $\{d_{ct},d_{pt},v_{ct}, v_{pt}, a_{ct}, a_{pt},d_{ct-1},d_{pt-1},v_{ct-1}, v_{pt-1}, a_{ct-1},$ $ a_{pt-1}\}$ including 12 different variables. After receiving the control commands, both players will move to their next position by some rules set in the environment. The state shifting is defined by a dynamic transition kernel $P(s_{t+1}|s_t, \mathbf{a}_t)$, which tells the probability players will observe a certain state $s_{t+1}$ by taking combination actions $\mathbf{a}_t=[a_{t,i},a_{t,-i}]$ in state $s_{t}$. In this paper, we take the car's (only has one type) point of view.

\subsection{Meta-reinforcement Learning}\label{ap:meta-RL}
Denote the state and the control inputs according to step $t$ by $s_t\in \mathcal{S}$, and $a_t\in \mathcal{A}$ in the environment that has a physical law $P_t(s_{t+1}|s_t,a_t)$, where $\mathcal{S}$ could be observed input from the sensor and $\mathcal{A}$ could be controlling output actions in a self-driving environment. The goal of meta-RL is to find an action choosing policy $\pi(a_t|s_t;\theta)$ embedded by weights $\theta$, where the policy $\pi$ can enlarge the expectation of reward $R(\tau):=\sum_{t=1}^H r(s_t,a_t)$ of trajectories $\tau:=(s_1,a_1,\ldots, s_H,a_H)$ with maximum $H$ steps. Suppose we have the probability that a trajectory occurs $q(\tau;\theta)$, then we can define the expected cumulative reward $J(\theta)= \mathbb{E}_{\tau\sim q(\cdot;\theta)}[R(\tau)]$ as the loss function.

For the meta-RL, the optimization problem is changed to:

\begin{align}
	\max_{\theta, \Phi} &\quad \mathbb{E}_{i\sim p} \mathbb{E}_{\mathcal{D}_i(\theta)\sim{q_i}}[J_i(\theta_i)] \label{eq:meta_opt}\\
	\st &\quad  \theta_i=\Phi(\theta,\mathcal{D}_i ), \nonumber
\end{align}

We introduce $i$ as the different tasks of meta-learning. $p$ is the meta-training distribution, and $p(i)$ denotes the probability that the agent is placed in the environment $i$ with different tasks in the meta-training. $\mathcal{D}_i$ is a batch of trajectories in the environment $i$, with the probability of each trajectory appearing being $q_i$. Considering the meta-RL model we choose in this work is the COLA~\cite{COLA} method, $\Phi(\theta,\mathcal{D}_i)$ will be obtained using the Lookahead Adaptation mechanism. Suppose environment mode or latent variable lives in the space $M$, the agent in this method forms its belief $b_t$ about the current environment mode, i.e., $b_t\in \Delta(M)$, by its history observation. After that, the agent conjectures that it is interacting with the stationary MDP for $L$ steps with probability $b(i)$. Therefore, we can get the trajectory segment $\tau^{L}:=(s_t,a_t,\ldots,s_{t+L-1}, a_{t+L-1}, s_{t+L})$ follows the distribution:
\begin{align}
    &q(\tau^L;b,\theta)=\\
    &\prod_{l=0}^{L-1}\!\pi(a_{t+l}|s_{t+l};\theta)\!\cdot\!\nonumber \prod_{l=0}^{L-1}\!\left(\sum_{i\in M}b(i)P_i(s_{t+l+1}|s_{t+l},a_{t+l})\!\right)\!.
\end{align}

In order to maximize its forecast of the future performance in $L$ steps, the adapted policy $\theta_t=\Phi_t(\theta)$ should maximize the forecast future performance: $ \max_{\theta'\in \Theta}\mathbb{E}_{q(\tau_t^L;b,\theta')}\sum_{l=0}^{L-1} r(s_{t+l},a_{t+l})$.

Unlike the reinforcement learning training, the agent has no access to the distribution $q(\cdot;b,\theta')$ in the online setting and hence, can not use policy gradient methods to solve for the maximizer. Following the approximation idea in trust region policy optimization (TRPO)~\cite{schulman15trpo}, the maximization of the future can be reformulated as  

\begin{align}\label{eq:trpo}
		\max_{\theta'\in\Theta} \quad &\mathbb{E}_{q(\cdot;b,\theta)}\nonumber
		\left[\prod_{l=0}^{L-1}\frac{\pi(a_{t+l}|s_{t+l};\theta')}{\pi(a_{t+l}|s_{t+l};\theta)}\sum_{l=0}^{L-1} r(s_{t+l},a_{t+l})\right]\nonumber\\
	&\st \quad \mathbb{E}_{s\sim q} D_{KL}(\pi(\cdot|s;\theta), \pi(\cdot|s;\theta'))\leq \delta,
\end{align}

where $D_{KL}$ is the Kullback-Leibler divergence. In the KL divergence constraint, we slightly abuse the notation $q(\cdot)$ to denote the discounted state visiting frequency $s\sim q$. The intuition is that when $\theta'$ is close to the base policy $\theta$ in terms of KL divergence, the estimated objective in \eqref{eq:trpo} using sample trajectories under $\theta$ returns a good approximation to $\mathbb{E}_{q(\tau^L;b,\theta')}\sum_{l=0}^{L-1} r(s_{t+l},a_{t+l})$. 

\subsection{Level-$k$ Thinking}\label{ap:klevel}
Level-$k$ thinking as a cognitive hierarchy theory, which is widely applied in multiplayer games, allows the level-$k$ thinking players to choose their best responses based on the assumption that all other players are level-$(k-1)$ thinkers. Denote $a^k$ as the action of level-$k$ player and $\mathbf{a}^{k-1}$ as all other $(k-1)$-th players' actions. Suppose we have a function $BR$ as a correspondence maps from an action to the set of its best responses. We could write the logic of level-$k$ thinking by:
\begin{align}
	a^{k}=BR(a^{k-1})
\end{align}

The relative clarity motivation of using level-$k$ thinking in our model is that it corresponds to reality's natural human thinking pattern. It makes the model more explicable and rational. Further, using level-k thinking in our method realizes we only need to know the player's intention, i.e., the conjecture or the utility function. Therefore, our structure is independent of understanding the complete rationality strategy, allowing the model to adapt online according to its conjectures.

\section{Proposed Mechanism}\label{sec:mechanism}
We propose this LK-MRL method to solve the vehicle-human interactions with the COLA algorithm and level-$k$ thinking strategy. To solve the optimization problem, we first give the measurement of performance for players $i,j$, which is $r_t:=r_{i,j}(s_t,\mathbf{a}_t)$. We define $\mathcal{O}_t=\{s_t,r_{t-1}\}$ be the set of agent's observations at time $t$, referred to as the information structure \cite{tao_info}. Then, the agent's policy $\{\pi_i:\mathcal{O}\to\Delta(\mathcal{A}_i)\}$ is a mapping from the past observations $\cup_{k=1}^t \mathcal{O}_k$ to a distribution over the action set $\Delta(\mathcal{A}_i)$ which is the probability simplex in $\mathbb{R}^{\mathcal{A}_i}$, for finite action set $\mathcal{A}_i$. For the optimization and controlling of the policy, assume the policy is generated by some machine learning model, e.g., neural networks, we define $\theta^k_{i,j}$ as the policy parameters of level-$k$ thinking player $i$ with type $j$. Since we are using the level-$k$ thinking method, we denote all superscript $k\in\mathbb{R}$ as the index of it belongs to the level-$k$ thinker.

\begin{figure*}[h]
    \centering
   \includegraphics[width=1.0\textwidth]{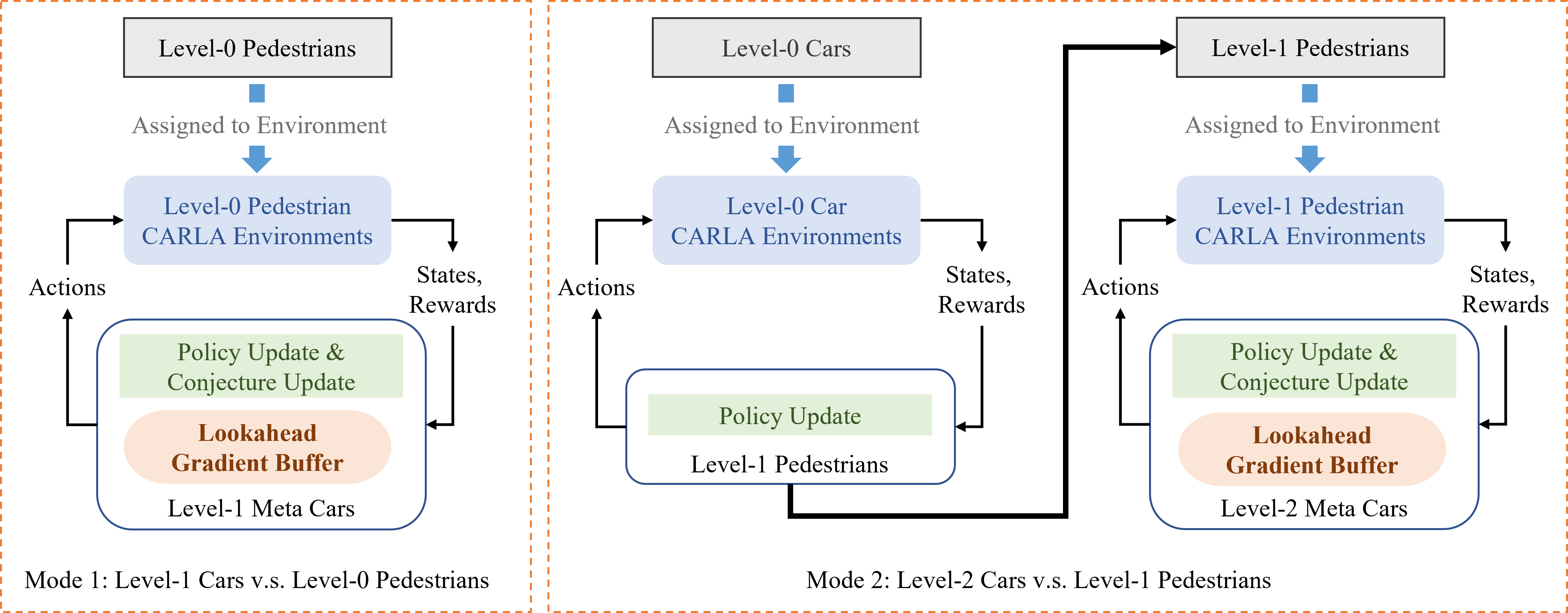}
    \caption{An illustration of the LK-MRL model is shown in the plot. The left-hand side is mode 1, which includes a level-$1$ car interacting with level-$0$ pedestrians. And the right-hand side is mode 2, which allows a level-$2$ car to interact with level-$1$ pedestrians. To obtain the final result of mode 2, we first need to process a level-$0$ car interacting with a level-$1$ pedestrians scenario, then assign the trained level-$1$ pedestrians to level-$1$ pedestrian CARLA environments for the training requirement of level-$2$ meta car.}
    \label{fig:LK-MRL}
\end{figure*}

We will first write the general case of the objective function and provide the optimizing algorithm.

\subsection{General Case of Model} \label{section:kmodel}
By the definition of level-$k$ thinking, the level-$k$ thinkers conjecture their opponents as level-(k-1) thinkers. To establish the model's foundation, let $V_{i,j}^k$ be the cumulative reward of level-$k$ thinking player $i$ with type $j$:
\begin{align}
    V_{i,j}^k&:=\E_{\pi_i,\pi_{-i}}\sum^H_{t=1}\!r_{i,j}\left[s_t,\!a_{t,i},\!a_{t,-i}\right].\nonumber\\
    a_{t,i}&\sim \pi_i(a_{t,i} | s_{t-1};\!\theta_{i,j}^k), a_{t,-i} \sim \pi_{-i}(a_{t,-i}| s_{t-1};\!\theta_{-i,j^{'}}^{k-1})
    \label{eq:value}
\end{align}
And then, we can write \eqref{eq:lgeneral} as a general form of the objective function $J_k$ for level-$k$ thinkers.

\begin{align}
    \max_{\theta} J^k_i:=\E_{j^{'}\in \Theta_{-i}}\E_{P(s_{t}|s_{t-1},\mathbf{a}_{t-1})}\!\left[V_{i,j}^k\!\right]\!.\label{eq:lgeneral}
\end{align}

\paragraph{Special case of level-$0$ thinker}
To simplify the training, we can use some constant policy agents as level-$0$ thinking players. However, we can also use the equation \eqref{eq:lgeneral} to obtain some RL policies for the level-$1$ thinkers and begin the game from level-$1$. Assuming the car and pedestrians all be level-$0$ thinkers, they think of themselves as the only player in this environment. Therefore, we could reduce the general form of the cumulative reward function to \eqref{eq:l0}.
\begin{align}
    V_{i,j}^0:=\E_{\pi_i}\sum^H_{t=1}r_{i,j}\left[s_t,\!\pi(a_{t,i}| s_{t-1};\!\theta_{i,j}^0)\right]\!.\label{eq:l0}
\end{align}
where $\mathbf{a}_t$ only contains $a_{t,i}$ since the level-$0$ player conjectures itself as the unique player in the environment. It is easy to see that both players will use a constant maximum strategy based on their reward functions. 

\paragraph{Case of higher level thinkers}
The higher level thinkers should follow the solution of \eqref{eq:lgeneral}. Suppose it is in the conventional logic, the level-$1$ interact level-$1$ game should be considered. However, with the LK-MRL model, we have a time-saving trick. If we have level-$0$ pedestrians, we can use it to obtain the level-$1$ car policy; If we have level-$1$ pedestrians, we can use it to obtain the level-$2$ car policy, and so on. It allows us to solve optimization problems alternately. Some researchers already show that normal humans usually only think about, at most, the second level of recursions in strategic games \cite{level-k}, thus we will not go further than the level-$2$ car policy, i.e., solve the optimization problem three times (the level-$0$ thinker's policy does not need calculation). In this paper, the experiment focuses on the level-$1$ car interact level-$0$ pedestrians situation, which is strong enough to support our work. Figure \ref{fig:LK-MRL} shows two modes of the LK-MRL model structure.

\subsection{Optimizing Algorithm}
This section is to find an optimizing algorithm for the constructed general objective function in section \ref{section:kmodel}. It is straightforward to notice that the objective function, no matter in which level of think, can be divided into two parts---optimizing pedestrians and optimizing cars. In both cases, we can assume their opponent uses a static model because the level-(k-1) thinker's strategy should be prior knowledge of the level-$k$ thinker.

For a level-$k$ ($k>0$) thinking car, the strategy it wants is optimal for all three types of pedestrians. Following the COLA method mentioned in \ref{ap:meta-RL}, we first obtain the car's base model shown in algorithm \ref{algo:rlc}.
\begin{algorithm}[h]	
\caption{RL-base}\label{algo:rlc}
	\begin{algorithmic}
		\State  \textbf{Input} Initialization $\theta_0$ in level $k$, $t=0$, step size $\alpha$, and the type of pedestrian $j$.
		\While{not converge}
		\State Sample a batch of trajectories $\mathcal{D}^{k-1}_{-i,j}$;
		\State $\theta_{h+1}=\theta_h+\alpha \hat{\nabla} J_i(\theta_h,\mathcal{D}^{k-1}_{-i,j})$ ;
		\State $h=h+1$.
		\EndWhile
		Let $\boldsymbol{\theta}_{i,j}^{k} = \theta_{h+1}$;
		\State\textbf{Return } $\boldsymbol{\theta}_{i,j}^{k}$
	\end{algorithmic}
\end{algorithm}

The COLA model uses its belief mechanism to make human-type conjectures. We can use many ways to realize it, e.g., latent type estimation \cite{pmlr-v97-rakelly19a} or a simple classification neural network like \cite{resnet} for image input. Suppose an inference network $\mathcal{F}$, as an approximate to some behavior sequences, is obtained by some machine learning algorithm. Suppose we are in step $t$. Since type is not directly observable to the vehicle, it is a latent variable to be estimated from the online observations. Based on it, we could claim the human-type conjecture $P_i=\mathcal{F}(s_1, a_{1,c}, a_{1,p};\cdots; s_t,a_{t,c},a_{t,p})$ is the probability of the vehicle's current opponent pedestrian's type. The function could be the belief of our policy to recognize which type of pedestrian it meets and make the Lookahead Adaptation on it. The pseudo-code is shown in algorithm \ref{algo:colac}.

\begin{algorithm}[h]	
	\caption{COLA}\label{algo:colac}
		\begin{algorithmic}
			\State  \textbf{Input} The base policy $\theta_1=\boldsymbol{\theta}_{i,j}^{k}$, human-type conjecture $P_i$, and gradient buffer $\mathcal{B}$, gradient sample batch size $D$ lookahead horizon length $L$ and $b_t$ is the belief related to $P_i$ of step $t$,  step size $\alpha$.
			\For {$t\in \{1,2,\ldots, H\}$}
			\State Obtain the state input $s_t$ from the environment;
			\State Implement the action using $\pi_{v}(a_{t,i}|s_t;\theta_t)$;
			\State Obtain the probability output from $P_i$;
			\If {$t \mod L = 0$ }
			\State Update the belief $b_t$;
			\State Sample $D$ gradients under different pedestrians' types from $\mathcal{B}$ according to $b_t$;
			\State $\theta_{t+1}=\theta_t+\alpha \cdot$ gradients sample mean ;
			\Else 
			\State $\theta_{t+1}=\theta_t$;
			\EndIf
			\EndFor
		Let $\theta_{i,j}^{k} = \theta_{t+1}$;
		\State\textbf{Return } $\theta_{i,j}^{k}$
		\end{algorithmic}
	\end{algorithm}
	
In order to show the optimal performance of the policy, we use the true type of the pedestrian as the output of human-type conjecture $P_i$.

\section{Simulation studies}\label{sec:simulation}
In this section, we first give the simulating context of this paper. Then, we divide our experiments into two kinds---mode 1 and mode 2. As we define in figure \ref{fig:LK-MRL}, mode 1 is a level-$1$ car interacting level-$0$ pedestrian scenario, which is more concentrated on the analysis of the model performance in each hierarchy of level-$k$ thinker. And mode 2 focuses on proving the retentivity ability for level-$k$ thinking properties of LK-MRL  by simulating a level-$1$ pedestrian and a level-$2$ car.

\begin{figure}[h]
	\centering
	\subfigure[Three types of pedestrian step mean speed plot.]{
		\begin{minipage}[b]{0.5\textwidth}
			\includegraphics[width=0.9\textwidth]{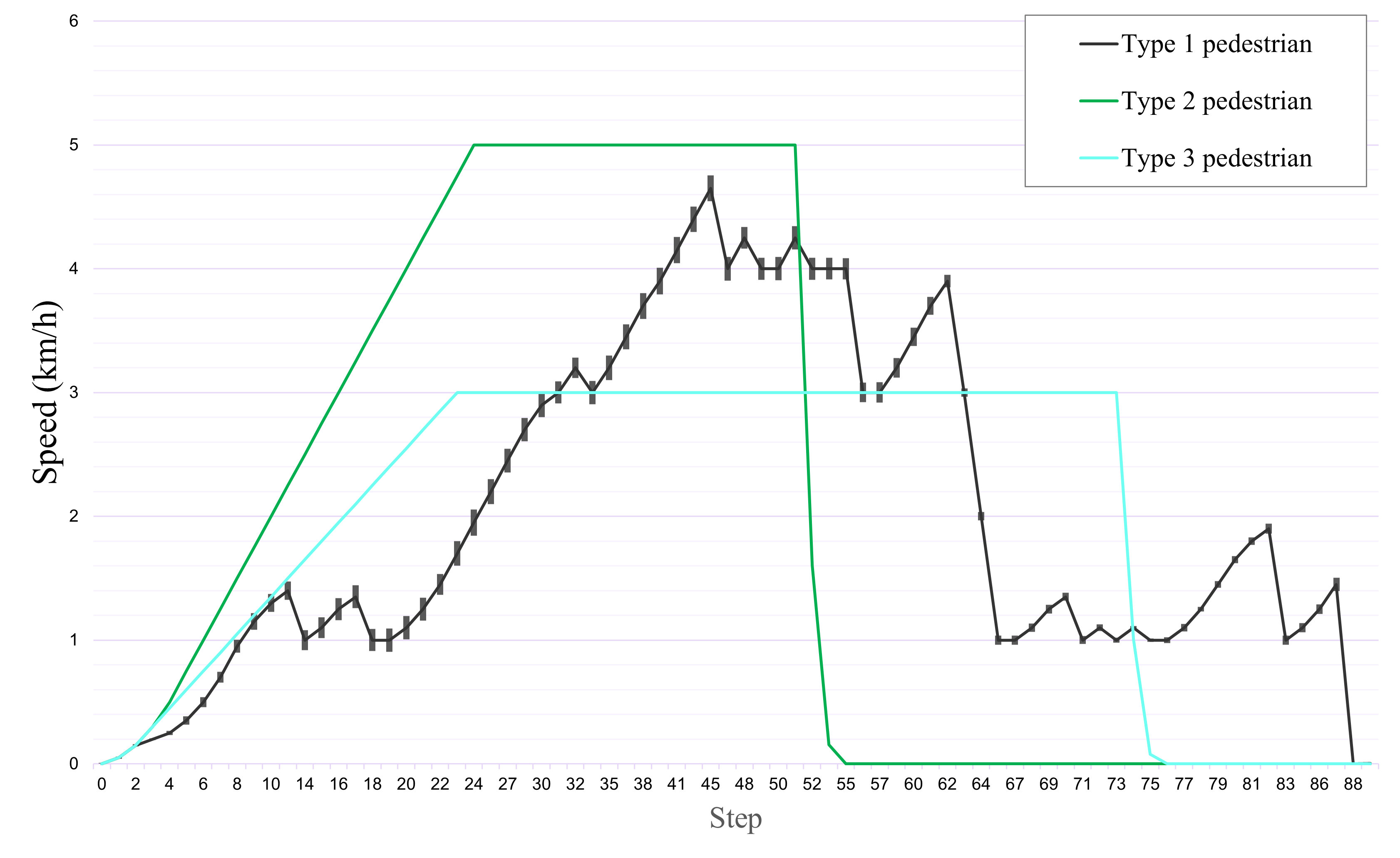}
		\end{minipage}
		\label{fig:P0-speed}
	}
	    \centering
    	\subfigure[Experimental Results From the mode 1 scenario: the (left) is about the mean and variance of episode reward and the collision rate; the (right) side is about the mean step speed performance.]{
    		\begin{minipage}[b]{0.5\textwidth}
   		 	\includegraphics[width=0.4\textwidth]{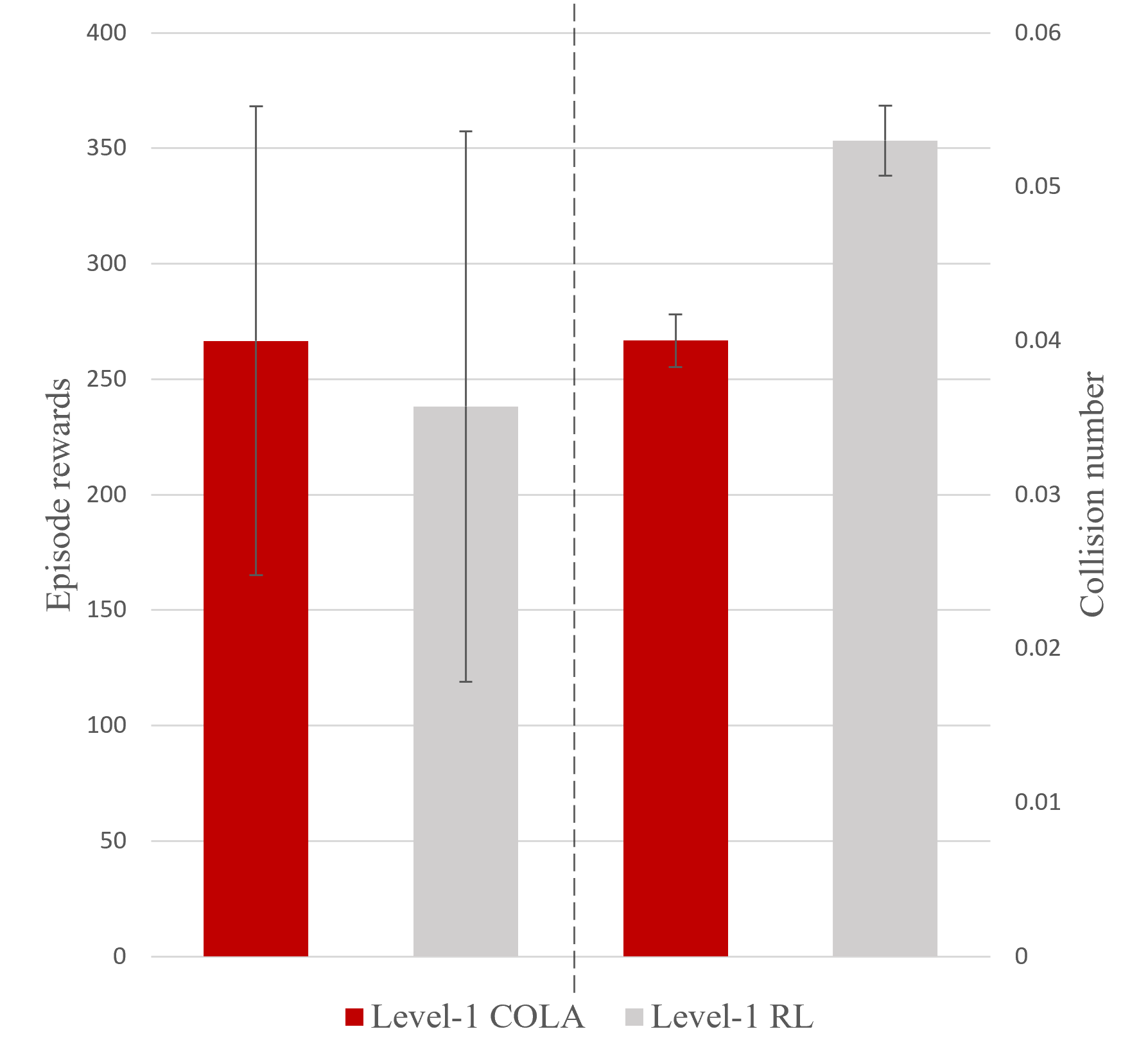}
		 	\includegraphics[width=0.5\textwidth]{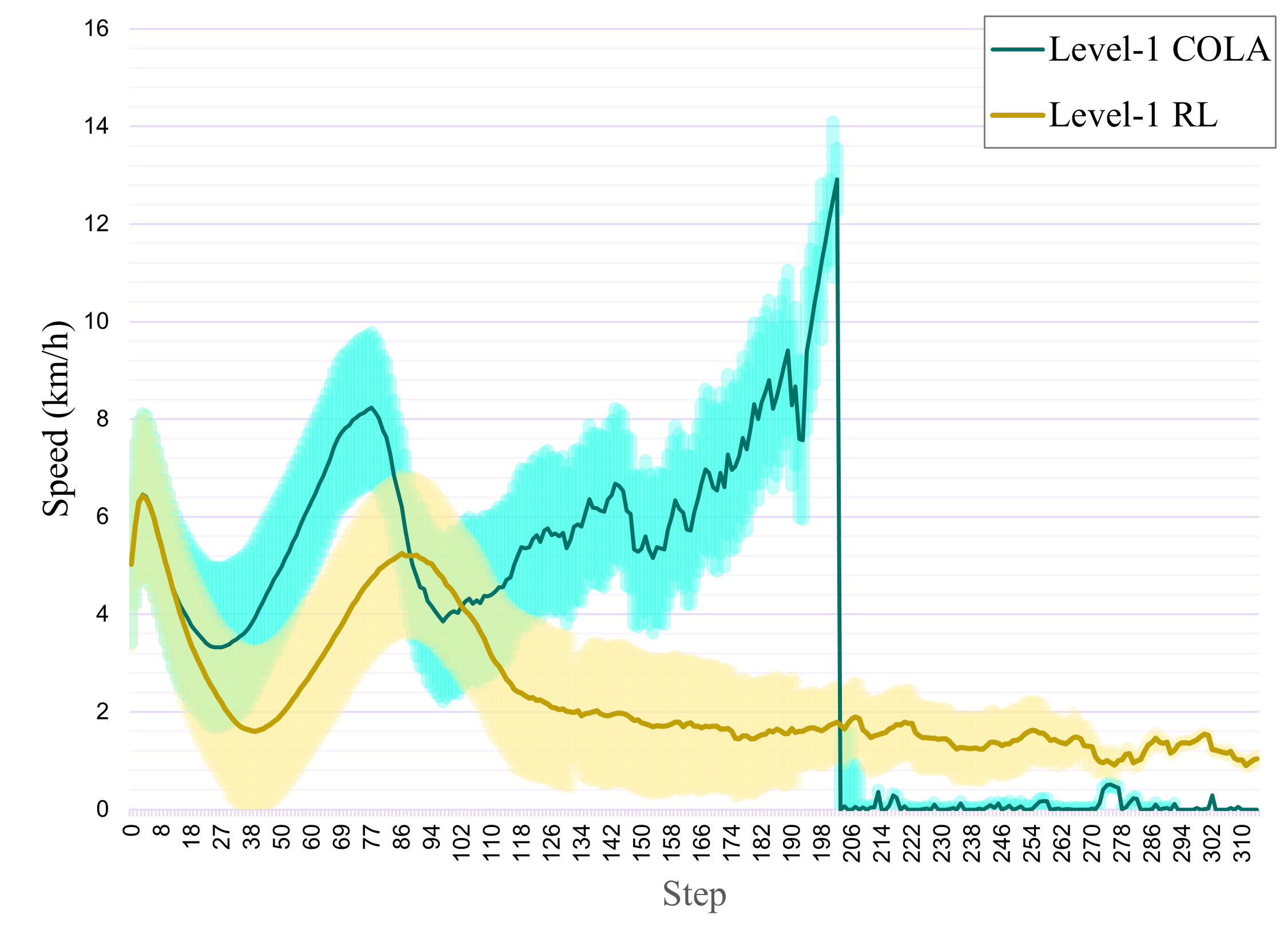}
    		\end{minipage}
		\label{fig:mode1results}
    	}
	\caption{In \ref{fig:P0-speed}, only type 1 pedestrian has variance since it involves randomness; In the (left) of \ref{fig:mode1results}, the COLA policy has a higher mean episode reward with minor variance, and a lower collision rate; In the (right) side of \ref{fig:mode1results}, we can see the speed performance of the COLA agent is also better (since the COLA agent will get to the destination with a shorter time, its mean step speed will decrease to zero earlier than RL agent).}
	\label{fig:mode1graphs}
\end{figure}
Our experiments use CARLA-0.9.4 \cite{carla} as the autonomous driving simulator. On top of the CARLA, we modify the API: Multi-Agent Connected Autonomous Driving (MACAD) Gym~\cite{macad-gym} to facilitate communications between learning algorithms and environments. The available actions for pedestrians are acceleration, deceleration, and cruising at the current speed, and for the car, throttle up, brake, and cruising at the current speed. To be more relevant to the actual traffic situation, we define three different types of pedestrians as basic level-$0$ thinkers, including (1) a pedestrian who moves randomly with a probability of 0.2 maintain the speed, 0.43 move acceleration, and 0.37 move backward; (2) a pedestrian accelerates to speed 5 km/h and maintain; (3) a pedestrian who accelerates to speed 3 km/h and maintain.

The Asynchronous Advantage Actor-Critic algorithm (A3C) with Adam optimizer mentioned in~\cite{Palanisamy:2018:HIA:3285236} for both pedestrians and cars are two $12\times 64\times 32\times 3$ neural networks with the rectified linear unit (ReLU) as its activation function. For pedestrians and the base policy of the car, the learning rate begins at $1\times 10^{-4}$, and the policy gradient update is performed end of every episode. The entropy regularized method \cite{entropy_bonus} is used. Once episode rewards stabilize, the learning rate will be changed to $1\times 10^{-5}$ and $1\times 10^{-6}$. The gradient buff of the COLA method will collect the gradient for every 10 steps and store 1000 episodic data for each type of pedestrian.

\subsection{Mode 1: level-$1$ car v.s. level-$0$ pedestrians}
We introduce mode 1, as mentioned in figure \ref{fig:LK-MRL}. It includes a level-$1$ car interacting with level-$0$ pedestrians. In this setting, we only use constant policy described in section \ref{sec:model_description} includes: (1) Randomly select with probability 0.2 maintain the speed, 0.43 move acceleration and 0.37 move backward; (2) Accelerating to 5 km/h and maintain; (3) Accelerating to 3 km/h and maintain. Figure \ref{fig:P0-speed} shows their speed patterns.

We compared the level-$1$ COLA agent (using learning rate $1\times10^{-3}$ and gradient buffer size equals 500 updates every 10 steps) with a vanilla A3C RL agent trained in the same environment. In our experiments, we give the COLA agent the type of pedestrians it meets since we want to test its optimal performance and reduce the complexity of model construction. The experimental results are summarized in figure \ref{fig:mode1results}. As shown in the picture, the COLA model performs better since it has a lower accidental rate with higher mean speed and shorter time spent.

\begin{figure}[h]
	\centering
	\subfigure[Step mean speed plot of type 1 of level-$0$ pedestrian, and level-$1$ pedestrian interact with three kinds level of car.]{
		\begin{minipage}[b]{0.5\textwidth}
			\includegraphics[width=0.9\textwidth]{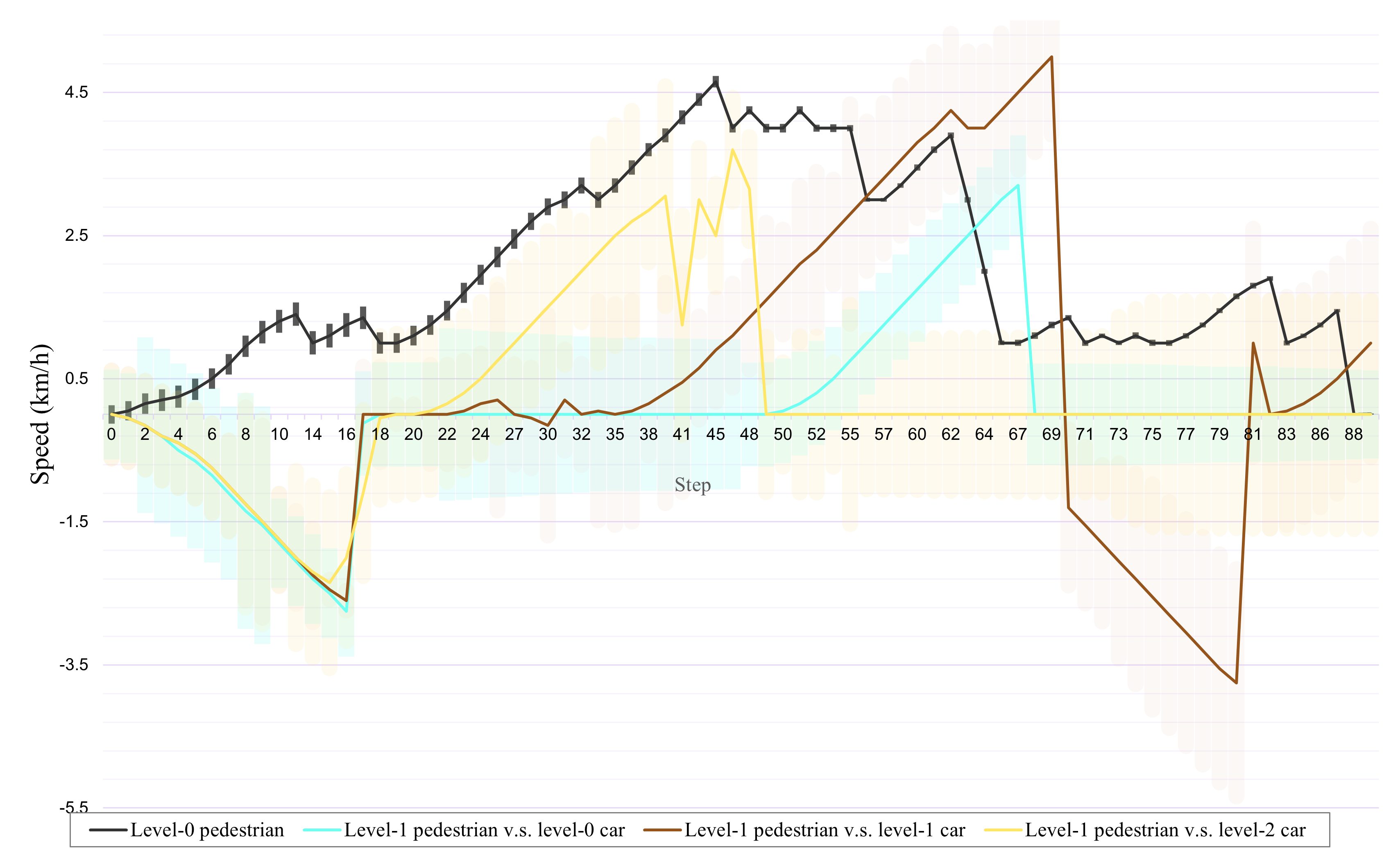}
		\end{minipage}
		\label{fig:P1-speed}
	}
	    \centering
    	\subfigure[Experimental Results From the mode 2 scenario: the (left) is about the mean and variance of episode reward and the collision rate; the (right) side is about the mean step speed performance.]{
    		\begin{minipage}[b]{0.5\textwidth}
   		 	\includegraphics[width=0.37\textwidth]{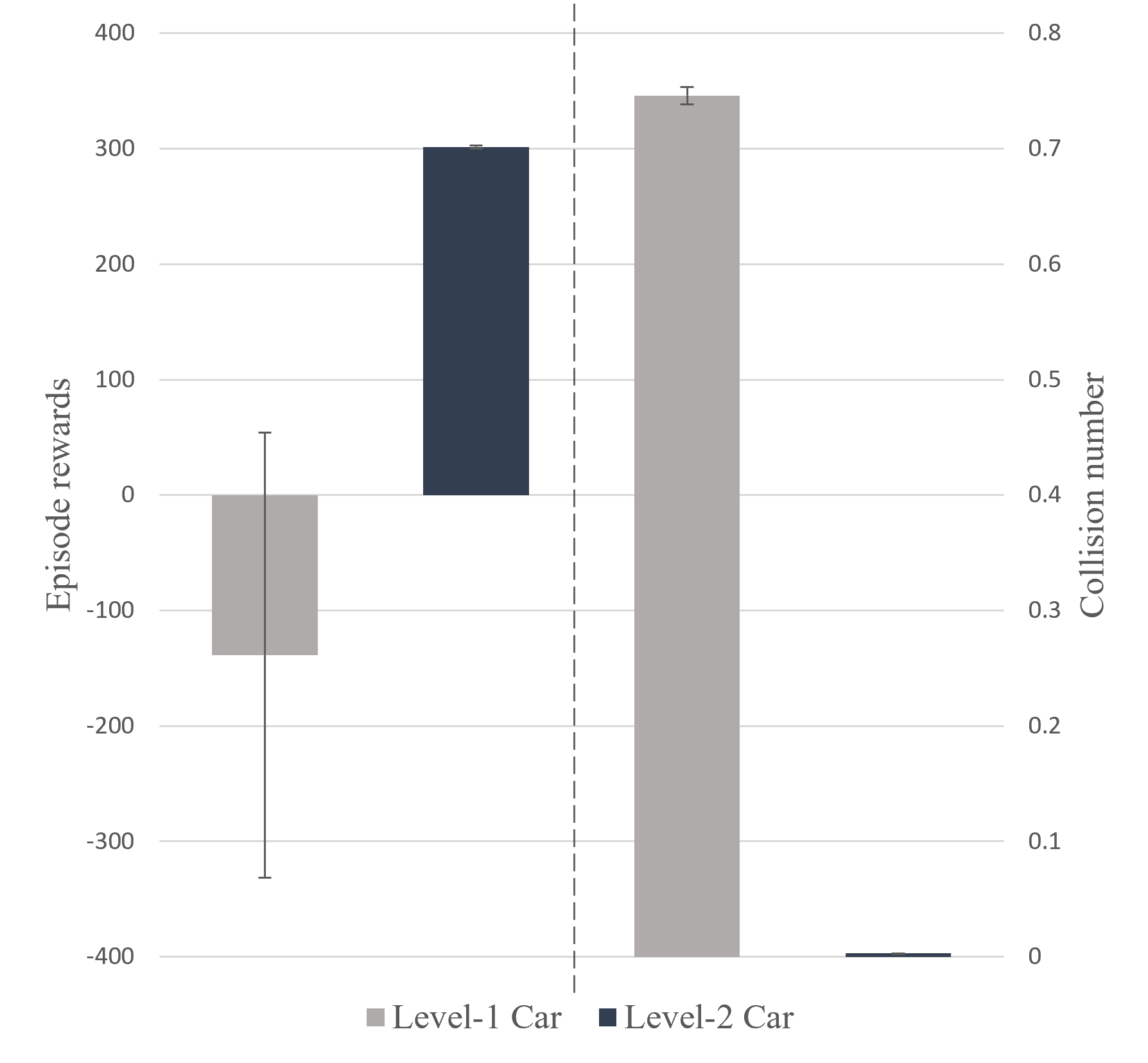}
		 	\includegraphics[width=0.55\textwidth]{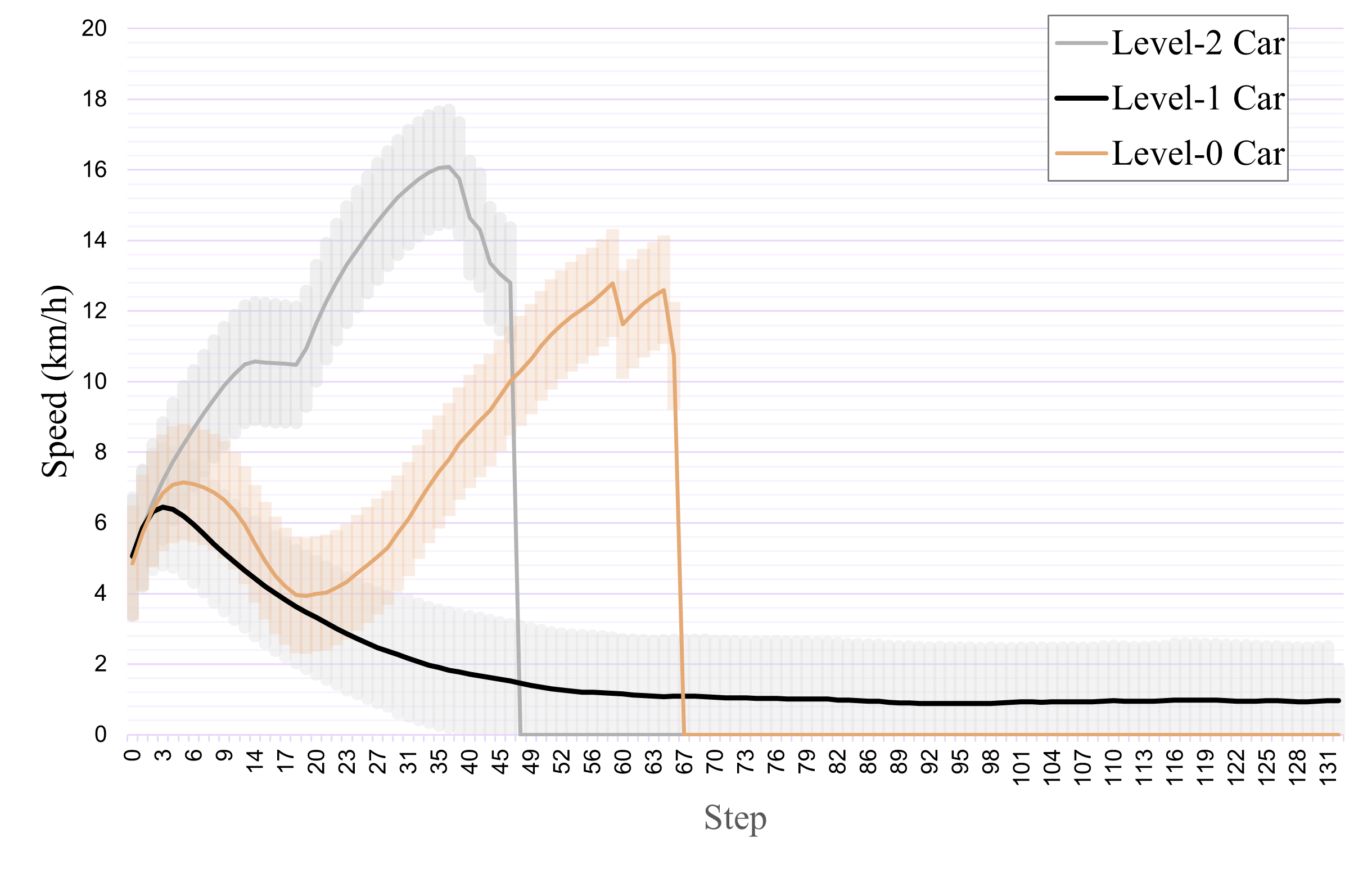}
    		\end{minipage}
		\label{fig:mode2results}
    	}
	\caption{In \ref{fig:P1-speed}, the level-$0$ pedestrian has a constant speed pattern with no response with the car, while the level-$1$ pedestrian has three different speed patterns; In the (left) of \ref{fig:mode2results}, the level-$2$ policy has a higher mean episode reward with minor variance, and a significant lower collision rate; In the (right) of \ref{fig:mode2results}, we can see the speed performance of the level-$2$ car agent is better (similar with the COLA agent, the level-$2$ agent will get to the destination with a shorter time).}
	\label{fig:mode2graphs}
\end{figure}

\subsection{Mode 2: level-$2$ car v.s. level-$1$ pedestrians}

In mode 2 experiments, we want to show the LK-MRL structure's retentivity ability for level-$k$ thinking properties. We first train a level-$1$ pedestrian with a given reward function based on a level-$0$ car with a given constant policy. The mean and variance of step speed are shown in figure \ref{fig:mode2results}.

Unlike constant level-$0$ pedestrians, the trained level-$1$ pedestrian performs further confrontations with different levels of cars. Its speed patterns shown in figure \ref{fig:P1-speed} are various, which means it can react with varying car actions. And since the level-$1$ car has the same level as the level-$1$ pedestrian, the result obtained for this situation is much worse than in mode 1. However, with 1000 episodes of fine-tuning training with a learning rate $1\times10^{-4}$, we get a prototype level-$2$ car, which shows significant performance improvement in the right-hand side of figure \ref{fig:mode2results}. As a result, we obtain a pedestrian that can think as level-$1$ thinker by conjecture.

\section{Conclusion}\label{sec:conclusion}
This work proposes a level-$k$ meta reinforcement learning (LK-MRL) structure based on the combination of speculative online lookahead adaptation (COLA) and hierarchical model works in urban traffic environments where vehicles need to interact with a diverse population of pedestrians. LK-MRL provides a solution by reducing the multi-agent question to alternate single-agent optimization and showing its ability for retentive level-$k$ thinking properties with the use of reinforcement learning.

\bibliographystyle{IEEEtran}
\bibliography{IEEEabrv,reference}
\vspace{12pt}

\end{document}